\documentclass[journal]{IEEEtran}

\usepackage{color}
\usepackage[square,numbers]{natbib}
\usepackage{amssymb}
\usepackage{amsmath}
\usepackage{amsthm}
\usepackage{epsfig}
\usepackage{eurosym}
\usepackage{graphicx}
\usepackage{graphics}
\usepackage{float}
\usepackage{subcaption}
\usepackage{xspace}
\usepackage{makecell}
\usepackage{multirow}
\usepackage{booktabs}
\usepackage[hidelinks]{hyperref}

\graphicspath{ {./images/} }

\def\BibTeX{{\rm B\kern-.05em{\sc i\kern-.025em b}\kern-.08em
    T\kern-.1667em\lower.7ex\hbox{E}\kern-.125emX}}

\begin{document}
\title{Probabilistic Prediction Markets with Intermittent Contributions}
\author{
    Michael~Vitali$^{1}$, Pierre~Pinson$^{1, 2, 3, 4}$\\[1em]
    \small $^{1}$ \textit{Dyson School of Design Engineering, Imperial College London, London, United Kingdom} \\
    \small $^{2}$ \textit{Halfspace, Denmark} \\
    \small $^{3}$ \textit{Department of Technology, Management and Economics, Technical University of Denmark, Denmark} \\
    \small $^{4}$ \textit{CoRE, Aarhus University, Denmark} \\[0.5em]
    \texttt{\small m.vitali24@imperial.ac.uk, p.pinson@imperial.ac.uk}
}
\maketitle

\begin{abstract}
Although both data availability and the demand for accurate forecasts are increasing, collaboration between stakeholders is often constrained by data ownership and competitive interests. In contrast to recent proposals within cooperative game-theoretical frameworks, we place ourselves in a more general framework, based on prediction markets. There, independent agents trade forecasts of uncertain future events in exchange for rewards. We introduce and analyse a prediction market operating with a fixed pool of agents that (i) accounts for the historical performance, (ii) adapts to time-varying conditions, while (iii) accommodates intermittent forecast submissions. The proposed design employs robust regression models to learn the optimal forecasts' combination whilst handling missing submissions. Moreover, we introduce a pay-off allocation mechanism that considers both in-sample and out-of-sample performance while satisfying several desirable economic properties. Case studies using simulated and real-world data demonstrate the effectiveness and adaptability of the proposed market design.
\end{abstract}

\section{Introduction}

Growing concerns about climate change and energy independence have led to a rapid expansion of solar and wind energy production. Forecasting plays a crucial role in decision-making processes, requiring high-quality data and advanced models. The increase in data availability represents an opportunity to refine these techniques and increase forecast quality, enabling companies to increase profits or reduce costs. However, data ownership issues often hinder progress, as companies are reluctant to share information and to collaborate due to competitive interests and security concerns.

A solution to increase collaboration is to give incentives in exchange for these data. This can be done in different ways, depending on whether the companies are willing to share data. If this is the case, different methods have been proposed for renewable power forecasting in the literature exploiting spatio-temporal relations. For solar power forecasting, a model incorporating exogenous data from different sources was developed \cite{Bessa2015}, while a similar approach has been applied to both wind and solar power forecasting \cite{Andrede2017}. In wind power forecasting, sparse spatio-temporal models were introduced in \cite{Dowell2016} and later extended in an online learning context \cite{Messner2019}.
Furthermore, to avoid the loss of privacy companies could face, privacy-preserving solutions using distributed learning methods have been proposed \cite{Cavalcante2017}, with further advancements addressing online settings \cite{Sommer2021, Goncalves2021}. However, all these solutions assume that the agents are willing to collaborate to improve forecasts and that they act rationally and truthfully. This is not always the case in practice. An alternative approach is to view the problem in a more general framework, through so called \emph{prediction markets}, in which agents choose to share their individual forecasts and get rewarded for their contribution to the resulting aggregate forecast.

In this context, prediction markets have been increasingly studied in the last decade, gaining popularity across various fields \cite{horn2014}, with two types of markets being proposed: {\it (i)} winner-takes-all markets, where only the best forecast is rewarded \cite{Witkowski2021}, and {\it (ii)} markets in which all participants are rewarded, since forecasts that are not the best can still carry valuable information (e.g. \cite{Raja2024}).

Nevertheless, existing solutions fail to consider some essential aspects of real-world applications such as: {\it (i)} real-time implementation, {\it (ii)} historical performance of the participants, and {\it (iii)} the ability to accommodate intermittent participation \textcolor{black}{(in the sense that a fixed pool of participants may submit forecasts irregularly)}. The first two challenges can be addressed through an online learning framework, whilst the third one can be seen as a missing data problem. To adaptively predict when certain features are missing, the literature has proposed robust variants of linear regression \cite{bertsimas2025, stratigakos2025} and online imputation strategies \cite{gronvald2024}. \textcolor{black}{In parallel, a pay-off allocation mechanism must determine how rewards are distributed among sellers, ensuring that the rule satisfies key economic properties. Two approaches are common in the literature. Allocations based on Shapley values \cite{shapley1953value} reward each seller for the marginal contribution they add to the combined forecast (see, for example, \cite{Goncalves2021, Pinson2022}), with online versions enabling dynamic pay-off distribution \cite{raja2023}. Alternatively, competitive scoring rules and wagering mechanisms \cite{Kilgour2004, Lambert2008} reward each seller based on the accuracy of their individual forecast relative to the pool. While contribution-based allocations effectively capture the value of information, they are often difficult to reconcile with truthfulness while retaining budget balance. In contrast, allocations based on scoring rules guarantee these properties but reward accuracy rather than the contribution to the combined forecast. Furthermore, these approaches typically focus on in-sample pay-offs, failing to account for out-of-sample performance.}

\textcolor{black}{We design an online prediction market\footnote{Code and data at \url{https://github.com/MichaelVitali/prediction_markets}} that addresses these three challenges. This is done by building upon concepts and approaches from online learning and market design. Online learning updates the combination weights over time based on the sellers' performance, and robust regression adapts the combination when forecasts are missing. For the pay-off, we follow the scoring-rule approach, which guarantees truthfulness and budget balance, and combine in-sample and out-of-sample rewards.} Building upon \cite{Raja2024}, our market design accommodates different clients and sellers that interact via a market operator. This results in the following contributions:
    \begin{itemize}
        \item \textcolor{black}{The market operator optimally combines input forecasts from a fixed pool of sellers, while accommodating intermittent submissions.}
        This is done through the use of a robust linear regression model \cite{bertsimas2025} that is able to predict in presence of missing data. Additionally, \textcolor{black}{we extend this model to an online setting with combination weights constrained to the probability simplex. Since a direct extension of the unconstrained updates is not suitable in this case, we update both the weights and the correction matrix with the feasible displacement of a projected gradient step}.
        \item \textcolor{black}{We adapt the competitive scoring rule of \cite{Kilgour2004, Lambert2008} to a market with intermittent participation. The reward is a combination of two allocations: an in-sample allocation, which rewards long-term accuracy through an exponentially weighted average of past scores, and an out-of-sample allocation, which rewards the accuracy of each session's forecast.}
    \end{itemize}

The remainder of the manuscript is organized as follows: Section \ref{sec:preliminaries} introduces the main concepts of the prediction market design. Then, Section \ref{sec:methodology} describes its methodological components, from adaptive robust linear regression to pay-off allocation. It is followed by Section \ref{sec:evaluation}, showing different test cases and corresponding results. \textcolor{black}{Section~\ref{sec:discussions} discusses
the trade-offs of the proposed design and its limitations.} Finally, Section \ref{sec:conclusions} concludes the paper and offers perspectives for future work.

\section{Preliminaries} \label{sec:preliminaries}

\subsection{Prediction Markets}
Market-based analytics can be broadly categorized into data markets and information markets, depending on whether the exchanged good is raw data or derived information. We focus on prediction markets, a specific subset of information markets. In prediction markets, participants trade forecasts about uncertain future events. These markets combine individual predictions to generate a final forecast, communicated to the buyer. Contributors to prediction markets are rewarded according to a predefined allocation rule, which may be based on the accuracy of their forecasts or on their contribution to the combined forecast. This is done via a mechanism with formal mathematical guarantees concerning desirable economic properties such as budget balance, symmetry, etc.

\subsection{Market Setup}
Our market design involves the interaction of multiple clients and sellers, through a central market operator. The market operates over a discrete time horizon $t = 1,2,\dots,T$. At each time step, the market facilitates the exchange of nonparametric probabilistic forecasts for a continuous variable of interest $Y_{t+k}$, where $k$ represents the lead time. These forecasts are represented by a set of quantiles $\hat{q}_\tau$ for various quantile levels $\tau \in (0,1)$. We define the participant roles as follows:
    \begin{itemize}
        \item \textbf{client} $c_i$: individual who requires a forecast for a variable of interest $Y_{t+k}$ (e.g., wind power generation at a set of lead times). We assume that the client seeking the forecast lacks internal forecasting capabilities and relies entirely on crowd-sourcing to obtain the forecast inputs. \textcolor{black}{Because there is no initial baseline to measure against, we consider a simplified scenario where the client pays a fixed monetary amount $U_t$ for each received forecast, representing the client's willingness to pay for receiving the combined forecast at each market session. This amount is agreed between the client and the platform when the market is opened, before any forecast is submitted. Since it does not depend on information reported by the client once the forecasts are received, the client cannot reduce its payment by under-reporting its utility. However, this framework can naturally be extended to a more general case where the client's utility, $U_t$, is evaluated as a function of the forecast improvement relative to an internal baseline. In that case, the client could misreport the performance of its baseline, and safeguards such as auditing would be required.}

        \item \textbf{seller} $s_i$: forecaster willing to provide forecasts for the variable of interest and in the format required by the client. We denote the set of participating sellers as $\mathbf{S}=\{s_1,s_2,\dots,s_n\}$. \textcolor{black}{Sellers may submit forecasts intermittently, in which case missing submissions receive zero reward. In a practical implementation, the platform could impose a minimum participation requirement, such as allowing no more than 10\% of sessions to be missed over a predefined period. This policy is not enforced in the present mechanism or experiments.}
        
        \textcolor{black}{The seller role is intended for forecasters with valuable predictive information who may either be unable to trade directly because of limited capital, market access, or infrastructure, or may simply prefer not to trade. In particular, some forecasters may wish to monetize their information without taking market positions or being exposed to trading losses and imbalance costs. Selling forecasts therefore provides an alternative.}
        \item \textbf{market operator (MO)}: central entity responsible for managing the market through a centralized digital platform. The operator coordinates the exchange by publishing the forecasting tasks, collecting seller submissions, and generating the aggregated final forecast. Furthermore, once the realization $y_{t+k}$ is observed, the operator calculates the relative performance of each seller, collects the client's payment, and redistributes rewards according to a predefined allocation rule.
        
    \end{itemize}
For clarity and simplicity, the following sections focus on a single-client setting interacting with a multiple-seller pool, as this captures the fundamental dynamics of the proposed mechanism.

\subsection{Market Operation Overview}
\begin{figure*}[!t]
        \centering
    \includegraphics[width=.9\textwidth, trim=0cm 0cm 0cm 1cm, clip]{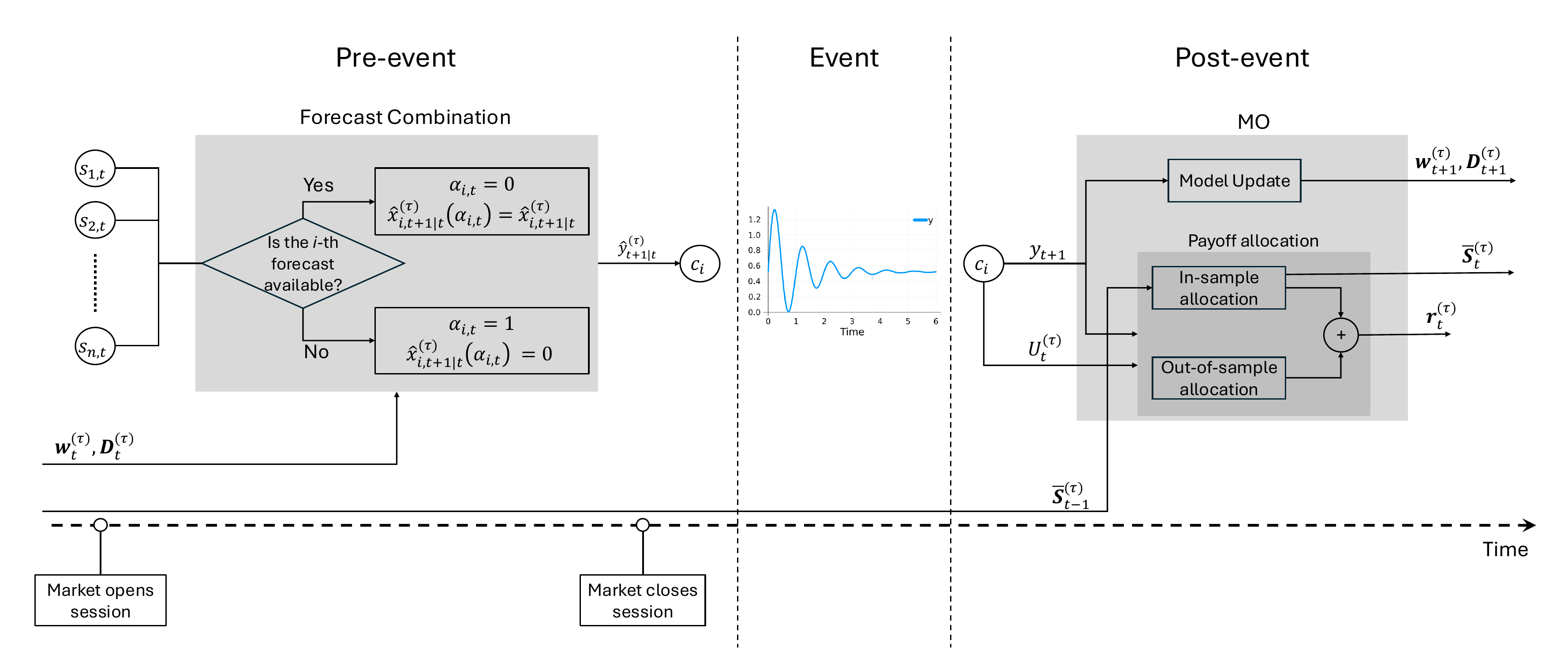}
        \caption{Market design overview}
        \label{fig:market-design}
        \vspace{-3mm}
    \end{figure*}
Fig. \ref{fig:market-design} provides an overview of the market and a high-level description of the sequential steps for the quantile level $\tau$ (and the example lead time $k = 1$). At the start of market operations, the absence of historical data presents a challenge for initializing both the ensemble weights and the pay-off mechanism. A practical solution is to initialize the weights $\mathbf{w}_t^{(\tau)}$ uniformly across all participating sellers, the correction matrix $\mathbf{D}^{(\tau)}$ at zero, and the long-term scores $\bar{\mathbf{S}}_{t}^{(\tau)}$ at zero so that sellers have to build their long-term score through accurate forecasts.

The process then begins with the market opening a session for a specified task. This session remains open for a limited time, during which sellers may submit their forecasts. If the $i$-th forecast is not submitted, the model finds the missing information from other forecasters (see Section~\ref{sec:adaptive_method} for details). Once the session closes, the market operator aggregates the submitted forecasts into a combined prediction and delivers it to the client. 

After the event occurs, the client reports the realization $y_{t+1}$ to the market operator and pays the agreed amount $U_t$. At this stage, the market operator performs two parallel operations: \textit{(1)} updating the model and propagating the new weights forward, and \textit{(2)} computing pay-off allocations so that each seller receives their corresponding reward. In addition, the updated in-sample values are carried over to the next session's pay-off calculation.

\textcolor{black}{The mechanism does not depend on a specific timing, which is set by the application. As an example, consider day-ahead wind power forecasting, where the client needs quantile forecasts for each hour of day $D+1$ before the day-ahead market gate closure at 12:00 on day $D$. The market operator opens a session at 00:00 on day $D$ and closes it at 10:00. The combined forecast is then computed within seconds and delivered to the client well ahead of gate closure. Since all realizations for day $D+1$ are available by 00:00 on day $D+2$, the operator has about ten hours to update the combination weights before the next session closes. All online operations, including the pay-off calculation, take only microseconds per session (see Table~\ref{tab:wall_clock}), so they fit easily within this window. If an update is not completed before a session closes, the most recent available parameters are used, and the new ones are applied in the following session.}

\subsection{Forecast Combination} \label{sec:prelim_forecast_combination}

We define the set of input forecasts as $\hat{\mathbf{X}}_{t} = \{ \hat{\mathbf{X}}_{1, t}, \hat{\mathbf{X}}_{2, t}, \dots, \hat{\mathbf{X}}_{n, t} \} \in \mathbb{R}^{n \times K \times |\mathbf{m}|}$, from the set of sellers $\mathbf{S}$, with $\mathbf{m}$ the set of quantiles and $K$ the maximum lead time. The forecast provided by the $i$-th seller is $\hat{\mathbf{X}}_{i, t} = \{ \hat{\mathbf{x}}_{i, t+1|t}, \dots, \hat{\mathbf{x}}_{i, t+K|t} \}$. The market operator generates the aggregated forecasts as a weighted average of input forecasts, for each quantile level $\tau$. The weight $w_{i}$ are learned online by minimizing the loss of the combined forecast.

As convention, we use $k$ for lead time, i.e., with $k=1,\hdots,K$. For simplicity, when describing methodological elements, we focus on the case of $k=1$ and a fixed quantile level $\tau$. The set of input forecasts then reduces to $\hat{\mathbf{x}}_t^{(\tau)} = \{ \hat{x}_{1, t+1|t}^{(\tau)}, \dots, \hat{x}_{n, t+1|t}^{(\tau)} \}$. The combined forecast is given as a convex combination of the input forecasts, i.e.,
    \begin{equation} \label{eq:forecast_combination_prel}
\hat{y}_{t+1|t}^{(\tau)} = \sum_{i=1}^{n} w_i \hat{x}_{i, t+1|t}^{(\tau)}, \quad  \sum_{i=1}^n w_i = 1, \quad w_i \geq 0
    \end{equation}

In our setup, the weights $w_i$'s are non-negative and sum to one, to improve interpretability. In the general case, such constraints are not strictly necessary: weights could be negative and not sum to one. Generally, the weights $w_i$ can be determined through various approaches, including simple equal weighting ($w_i = 1/n$) as a robust baseline, weights learned from past data (as for our case), or a wagering-based mechanism where weights are proportional to the sellers' wagers as a proxy for their self-assessed confidence \cite{Raja2024}. For a comprehensive overview of the state of the art with forecast combination, we refer the reader to \cite{wang2022}.

\subsection{Pay-off Allocation}

A pay-off function serves as the cornerstone of any market mechanism designed for data sharing or forecast combination. Its primary objective is to distribute the generated utility among the market participants (e.g.,  forecasters) commensurate with their individual performances. Carefully designing this function is critical to incentivize market participation, while ensuring that the mechanism remains economically viable and robust.
To achieve this, pay-off allocations must satisfy a set of rigorous economic properties \cite{Raja2024, Pinson2022}, which are formally detailed and proven later in this paper (see Section~\ref{properties}).

Beyond these foundational properties, the pay-off function is structurally divided into \textit{in-sample} and \textit{out-of-sample} allocations. Given that we operate in an online setting without a distinct training phase, the in-sample reward is specifically designed to reward \textit{consistency} and the participants' long-term accuracy.  Conversely, the out-of-sample allocation is utilized to reward \textit{instantaneous performance} (or, genuine forecast quality). This component is strictly tied to the realization of future events, ensuring that rewards are allocated based on the actual accuracy of a forecast at the moment it is provided.

By combining these two dimensions, the mechanism balances long-term reliability with immediate predictive excellence in a dynamic environment. In the following sections, we will describe how these rewards are calculated using specific scoring rules to ensure that the prediction market operates efficiently and remains incentive-compatible for all participants.

\section{Methodology} \label{sec:methodology}
\subsection{Bounded Report Space and Normalization}
\textcolor{black}{
Before performing the forecast combination and pay-off allocation, all forecasts and realizations are mapped to the common scale $[0, 1]$. We assume that the
variable of interest has known lower and upper bounds $\underline{y}<\overline{y}$. The market operator clips each value to this admissible interval and applies the min--max transformation
\begin{equation}
    g(v) = \frac{\operatorname{clip}(v,\underline{y},\overline{y})-\underline{y}}{\overline{y}-\underline{y}}.
    \label{eq:minmax_normalization}
\end{equation}
The same transformation is applied to all submitted forecasts and to the realization. The bounds are specified before market operation using physical or operational knowledge and are not estimated from the evaluation data.
}
\textcolor{black}{For notational simplicity, throughout this section, $y_{t+k}$,
$\hat{x}_{i,t+k|t}^{(\tau)}$, and $\hat{y}_{t+k|t}^{(\tau)}$ denote the normalized realization, normalized seller forecast, and normalized combined forecast, respectively. Hence, all forecast-combination updates and reward calculations are performed using values in $[0,1]$. Whenever a forecast in the original units is required, the market operator applies
\begin{equation}
    g^{-1}(v)=\underline{y} + (\overline{y}-\underline{y})v.
    \label{eq:inverse_minmax}
\end{equation}}

\vspace{-10mm}
\subsection{Forecast Combination} \label{sec:forecast_combination}
Forecast combination is performed using a linear regression framework, as in \eqref{eq:forecast_combination_prel}. To adapt to potential changes in market dynamics over time, the method is extended to an online learning setting, allowing the model to continuously track the optimal combination of forecasts (weights). For each quantile $\tau$ of interest a different linear regression is learned, enabling separate learning dynamics for each quantile level. \textcolor{black}{As a consequence, the combined quantiles may cross. The quantiles are therefore sorted at each time step following the monotone rearrangement in \cite{chernozhukov2010}, delivered to the client and used in the model updates. The sorting does not affect the sellers' payments, which depend only on their individual forecasts.}

The model weights are updated online using online gradient descent (OGD). To derive the update rule, we first define the objective function. Since we focus on nonparametric probabilistic forecasting, we employ the quantile loss (or, pinball loss), defined as
    \begin{equation}
        \mathcal{L}^{(\tau)} (y, \hat{y}) = 
        \begin{cases}
            (y - \hat{y}) \tau & \text{if } y \geq \hat{y} \\
            (\hat{y} - y) (1 - \tau) & \text{if }\hat{y} > y
        \end{cases}
    \end{equation}
where $y$ is the observation, $\hat{y}$ is the aggregated forecast, and $\tau$ is the target quantile level. The quantile loss is non-differentiable at zero, requiring us to employ the sub-gradient approach, e.g., studied in \cite{shen2025}.

Applying the chain rule to the loss function with respect to an individual weight $w_{i, t}$, we obtain the following sub-gradient:
    \begin{equation} \label{eq:loss_derivative_weights}
        \begin{split}
            &\nabla_{w_{i, t}} \mathcal{L} (y_{t+1}, \hat{y}_{t+1|t}^{(\tau)}) =\\
            &\qquad \begin{cases}
                -\tau \hat{x}_{i, t+1|t}^{(\tau)} & \text{if } y_{t+1} \geq \hat{y}_{t+1|t}^{(\tau)} \\
                (1 - \tau) \hat{x}_{i, t+1|t}^{(\tau)} & \text{if } \hat{y}_{t+1|t}^{(\tau)} > y_{t+1}
            \end{cases}
        \end{split}
    \end{equation}

Consequently, the weight vector is updated at each step with
    \begin{equation} \label{eq:single_step_update_linear}
        \mathbf{w}_{t+1}^{(\tau)} = \Pi_{\mathcal{H}} \left( \mathbf{w}_{t}^{(\tau)} - \eta \nabla_{\mathbf{w}_t} \mathcal{L}(y_{t+1}, \hat{y}_{t+1|t}^{(\tau)}) \right)
    \end{equation}
where $\eta$ is the learning rate and $\Pi_{\mathcal{H}}(\cdot)$ denotes the Euclidean projection onto the feasible set $\mathcal{H}$ defined by the constraints in \eqref{eq:forecast_combination_prel}. Specifically, $\mathcal{H}$ is the probability simplex, i.e.,
    \begin{equation} \label{eq:simplex}
        \mathcal{H} = \left\{ \mathbf{w} \in \mathbb{R}^n : \sum_{i=1}^n w_i = 1, w_i \geq 0 \right\}
    \end{equation}

While the optimization above is defined for a single lead time, operational forecasting often requires predicting across a multi-step horizon up to a maximum lead time $K$. To efficiently and stably optimize the aggregation weights across all $K$ lead times, we extend the OGD approach using mini-batch online gradient descent.
Specifically, we partition the full forecasting horizon into disjoint subsets. Let $B \subseteq \{1, \dots, K\}$ represent one such mini-batch, consisting of a subset (or potentially the entirety) of lead time indices $k$. Instead of performing an update for each individual lead time, the gradient is computed by averaging the individual sub-gradients over all lead times within the mini-batch $B$:
    \begin{equation} \label{eq:minibatch_grad}
        \nabla_{\mathbf{w}_t} \mathcal{L}_B = \frac{1}{|B|} \sum_{k \in B} \nabla_{\mathbf{w}_t} \mathcal{L}(y_{t+k}, \hat{y}_{t+k|t}^{(\tau)})
    \end{equation}
This batch-averaged gradient replaces the single-step gradient in the update rule defined in (\ref{eq:single_step_update_linear}).

\subsection{Adaptive Robust Linear Regression} \label{sec:adaptive_method}

The standard linear combination assumes that sellers submit predictions at every market session. However, this is unrealistic in real-world scenarios. Thus, \textcolor{black}{to handle intermittent submissions}, we adopt a robust variant of the linear regression (LR) model capable of handling missing forecasts. Originally introduced in \cite{bertsimas2025} and applied in \cite{stratigakos2025}, this method learns a linear correction matrix $\mathbf{D}^{(\tau)}$ among input forecasts and uses this correction to modify the combination weights when some input forecasts are unavailable. In essence, the model compensates for the missing information by extracting additional information from the remaining available forecasts. 
In this work, we extend this method to the online learning setting. \textcolor{black}{Since our combination weights are constrained, this extension is not just a matter of replacing the batch solver with the online updates, but also requires accounting for the projection.}

We model the availability of the forecasts using a binary variable $\boldsymbol{\alpha}_t \in \{0, 1\}^n$, which takes value 1 if the $i$-th forecast is unavailable at time $t$, and 0 otherwise. The forecast vector is redefined such that missing values are zeroed out,
    \begin{equation}
        \hat{x}_{i, t+1|t}^{(\tau)}(\alpha_{i, t}) = 
        \begin{cases}
            \hat{x}_{i, t+1|t}^{(\tau)} & \text{if } \alpha_{i, t} = 0 \\
            0 & \text{if } \alpha_{i, t} = 1
        \end{cases}
    \end{equation}

To compensate for missing forecasts, we dynamically adjust the weights using the linear correction matrix $\mathbf{D}^{(\tau)}_t$. Here, $\mathbf{w}^{(\tau)}_t$ represents the baseline model when all forecasts are available. Furthermore, we explicitly mask out the weights of unavailable forecasts using the Hadamard product
    \begin{equation}
        \tilde{\boldsymbol{\theta}}^{(\tau)}_t = (\mathbf{w}_t^{(\tau)} + \mathbf{D}_t^{(\tau)} \boldsymbol{\alpha}_t) \odot (1 - \boldsymbol{\alpha}_t)
    \end{equation}

To ensure the final aggregation is a valid convex combination, the projection step is applied exclusively to the available forecasters. Let $\mathcal{A}_t = \{i \mid \alpha_{i, t} = 0\}$ denote the set of active forecaster indices. We extract the weights corresponding to the active set, project them onto the probability simplex $\mathcal{H}$, and reconstruct the final weights vector $\boldsymbol{\theta}_t^{(\tau)}(\boldsymbol{\alpha}_t)$ by assigning zero weight to missing forecasts
    \begin{equation} \label{eq:projected_effective_weights}
        \theta_{i, t}^{(\tau)}(\alpha_{i, t}) = 
        \begin{cases} 
            \left[ \Pi_{\mathcal{H}}([\tilde{\boldsymbol{\theta}}_t^{(\tau)}]_{\mathcal{A}_t}) \right]_i & \text{if } i \in \mathcal{A}_t \\ 
            0 & \text{if } i \notin \mathcal{A}_t
        \end{cases}
    \end{equation}

The robust forecast combination is then computed as
    \begin{equation} \label{eq:robust_aggregation}
        \hat{y}_{t+1|t}^{(\tau)} = \boldsymbol{\theta}_t^{(\tau)}(\boldsymbol{\alpha}_t)^\top \hat{\mathbf{x}}_t^{(\tau)}(\boldsymbol{\alpha}_t)
    \end{equation}

\textcolor{black}{
The combined forecast in \eqref{eq:robust_aggregation} is defined as long as at least one seller submits a forecast, i.e., $|\mathcal{A}_t| \geq 1$. When only one seller is available, the projection in \eqref{eq:projected_effective_weights} assigns it a weight of one, and the combined forecast equals that seller's forecast. The mechanism does not enforce a minimum participation requirement or a maximum missing-data rate, although such rules could be added in a practical implementation.
}

\textcolor{black}{
We extend this method to the online learning setting by deriving update rules for both $\mathbf{w}$ and $\mathbf{D}$. Let $\nabla_{\theta_{i, t}}\mathcal{L}(y_{t+1}, \hat{y}_{t+1|t}^{(\tau)})$ denote the gradient of the pinball loss w.r.t. the weights used in  \eqref{eq:projected_effective_weights}, whose $i$-th entry is}
    \begin{equation}
        \begin{split}
            &\nabla_{\theta_{i, t}}\mathcal{L}(y_{t+1}, \hat{y}_{t+1|t}^{(\tau)}) = \\ 
            &\quad \begin{cases}
                - \tau \hat{x}_{i, t+1|t}^{(\tau)} (1 - \alpha_{i, t}) & \text{if } y_{t+1} \geq \hat{y}_{t+1|t}^{(\tau)} \\
                (1 - \tau) \hat{x}_{i, t+1|t}^{(\tau)} (1 - \alpha_{i, t}) & \text{if } \hat{y}_{t+1|t}^{(\tau)} > y_{t+1}
            \end{cases}
        \end{split}
    \end{equation}
    
\textcolor{black}{
A direct extension of the unconstrained updates in
\cite{bertsimas2025, stratigakos2025} would apply a gradient step to the
weights and the correction matrix.
However, this approach is not suitable when the combination weights are constrained to the probability simplex. The projection onto the simplex redistributes weight across components to enforce the sum-to-one constraint. As a result, a step that decreases the weight of an active seller can alter the weights of unavailable sellers, even though they do not participate in the current session. A second difficulty arises when the optimal prediction lies outside the convex hull of the active sellers' predictions. In that case, the loss gradient may continue to point beyond the feasible region. Applying this gradient directly at every update can therefore cause the correction matrix $\mathbf{D}^{(\tau)}$ to accumulate changes that cannot be realized by the combination weights, eventually degrading the base weights. These issues do not arise in \cite{bertsimas2025, stratigakos2025}, where the weights are not constrained to the simplex.
}

\textcolor{black}{
We instead compute the feasible displacement induced by the gradient step. First, we apply the step to the active combination weights and project the result onto the simplex. The difference between the projected and original active weights defines $\mathbf{d}_t^{(\tau)}$. This projected gradient step accounts for the simplex boundary while leaving the displacements of unavailable sellers at zero:
    \begin{equation}
        \boldsymbol{\theta}^{+}_t = \Pi_{\mathcal{H}} \left(
            [\boldsymbol{\theta}_t^{(\tau)}(\boldsymbol{\alpha}_t)]_{\mathcal{A}_t}
            - \eta [\nabla_{\theta_{i, t}}\mathcal{L}(y_{t+1}, \hat{y}_{t+1|t}^{(\tau)})]_{\mathcal{A}_t} \right)
            \label{eq:pg_step}
    \end{equation}
    \begin{equation}
        d_{i,t}^{(\tau)} =
        \begin{cases}
            \theta^{+}_{i,t} - \theta_{i,t}^{(\tau)}(\alpha_{i,t})
            & \text{if } i \in \mathcal{A}_t,\\
            0 & \text{if } i \notin \mathcal{A}_t,
        \end{cases}
        \label{eq:pg_displacement}
    \end{equation}
    \begin{equation}
        \mathbf{w}_{t+1}^{(\tau)} =
        \Pi_{\mathcal{H}}\left(
            \mathbf{w}_t^{(\tau)}+\mathbf{d}_t^{(\tau)}
        \right),
        \label{eq:update_robust_weights}
    \end{equation}
    \begin{equation}
        \mathbf{D}_{t+1}^{(\tau)} =
        \mathbf{D}_t^{(\tau)} + \mathbf{d}_t^{(\tau)}\boldsymbol{\alpha}_t^\top,
        \label{eq:update_linear_correction}
    \end{equation}
where $\eta$ is the learning rate. 
}
    
\subsection{Pay-off Allocation} \label{sec:pay-off_allocation}
\textcolor{black}{
When the client provides the true realization $y_{t+1}$, the total available reward $U_t$ has to be split according to both the in-sample and out-of-sample allocations. We assume that the total reward is divided equally among the quantile levels $U_t^{(\tau)} = U_t / |\mathbf{m}|$. For each quantile, a fraction
$\delta \in [0, 1]$ is assigned to the in-sample component and the remaining fraction $1-\delta$ to the out-of-sample component. The value of $\delta$ is agreed between the client and the market operator when the market is opened. We have that the reward for the $i$-th seller at time $t$ for the quantile level $\tau$ is
    \begin{equation}
        R_{i, t}^{(\tau)} = U_{t}^{(\tau)} [\delta r_{i, t}^{\text{is}(\tau)} + (1-\delta) r_{i, t}^{\text{oos}(\tau)}]
    \end{equation}
where $r_{i, t}^{\text{is}(\tau)}$ is the in-sample allocation and $r_{i, t}^{\text{oos}(\tau)}$ the out-of-sample one. Finally, the total reward for each seller is defined as $R_{i, t} = \sum_{\tau \in \mathbf{m}} R_{i, t}^{(\tau)}$.
}

\subsubsection{In-sample allocation}
\textcolor{black}{
The in-sample allocation rewards sellers who consistently provide accurate forecasts over time. Let $N_t = |\mathcal{A}_t|$ be the number of active sellers at time $t$. For each $i \in \mathcal{A}_t$, we define the normalized quantile score as
\begin{equation}
    sc_{i,t}^{(\tau)} =
    \begin{cases}
        1 - \dfrac{\mathcal{L}^{(\tau)}\big(y_{t+1}, \hat{x}_{i,t+1|t}^{(\tau)}\big)}{\mathcal{L}^{(\tau)}_{\text{max}}}, & i \in \mathcal{A}_t,\\[3mm]
        0, & i \notin \mathcal{A}_t,
    \end{cases}
    \label{eq:score}
\end{equation}
where $\mathcal{L}^{(\tau)}_{\text{max}} = \max\{\tau, 1-\tau\}$ is the maximum loss (since forecasts and realizations lie in $[0, 1]$) and $sc_{i,t}^{(\tau)} \in [0,1]$. When forecasts are issued for several lead times, the session score is the average of \eqref{eq:score} over all lead times $k = 1, \dots, K$, which also lies in $[0,1]$.
}

\textcolor{black}{
To capture long-term performance, the score is accumulated through an exponentially weighted average
\begin{equation}
    \bar{S}_{i,t}^{(\tau)} = \lambda \bar{S}_{i,t-1}^{(\tau)} + (1-\lambda)\, sc_{i,t}^{(\tau)},
    \label{eq:long_term_score}
\end{equation}
with $\bar{S}_{i,0}^{(\tau)} = 0$, where $\lambda$ is the forgetting factor which can be expressed through the effective window size $n_\lambda = (1-\lambda)^{-1}$ \cite{Pinson2022}. Since a missing seller receives a zero score, its long-term score decays while it is inactive, so that the in-sample allocation rewards both accuracy and consistent participation. Initialising $\bar{S}_{i,0}^{(\tau)} = 0$ means that sellers at the start of the market are not trusted a priori: their long-term score grows only as evidence of accurate forecasting accumulates, so that a few good sessions cannot produce a high long-term reward. Since $\bar{S}_{i,t}^{(\tau)}$ is a convex combination of scores in $[0,1]$ and the initial value $0$, it holds that $\bar{S}_{i,t}^{(\tau)} \in [0,1]$.
}

\textcolor{black}{
The in-sample utility is allocated using the competitive scoring rule of
\cite{Kilgour2004}, in the generalized form of \cite[Theorem~2]{Lambert2008}. Each active seller is initially assigned an equal share of the in-sample pot, which is then adjusted according to the difference between its long-term score and the average long-term score of the other active sellers
\begin{equation}
    \begin{aligned}
        & r_{i,t}^{\mathrm{is}(\tau)} = \\[-1mm]
        & \begin{cases}
            \displaystyle \frac{1}{N_t}\left[1+\bar{S}_{i,t}^{(\tau)}-\frac{1}{N_t-1}\sum_{\substack{j\in\mathcal{A}_t\\j\neq i}}\bar{S}_{j,t}^{(\tau)}\right], & i\in\mathcal{A}_t,\;N_t\geq2, \\[3mm]
            1, & i\in\mathcal{A}_t,\;N_t=1,\\
            0, & i\notin\mathcal{A}_t.
        \end{cases}
    \end{aligned}
    \label{eq:in_sample_reward}
\end{equation}
Thus, an active seller receives more than the equal allocation when its long-term score exceeds the average of the other active sellers, and less otherwise. Missing sellers receive zero. When only one seller is active, no relative performance comparison is possible and the full in-sample allocation
is assigned to that seller.
}

\subsubsection{Out-of-sample allocation}
\textcolor{black}{
The out-of-sample allocation rewards sellers for their accuracy in the current session, regardless of their past performance or participation. It applies the equal-stake rule in \eqref{eq:in_sample_reward} to the session score in \eqref{eq:score} instead of the long-term score
\begin{equation}
    \begin{aligned}
        & r_{i,t}^{\mathrm{oos}(\tau)} = \\[-1mm]
        & \begin{cases}
            \displaystyle \frac{1}{N_t}\left[1+sc_{i,t}^{(\tau)}-\frac{1}{N_t-1}\sum_{\substack{j\in\mathcal{A}_t\\j\neq i}}sc_{j,t}^{(\tau)}\right], & i\in\mathcal{A}_t,\;N_t\geq2, \\[3mm]
            1, & i\in\mathcal{A}_t,\;N_t=1,\\
            0, & i\notin\mathcal{A}_t.
        \end{cases}
    \end{aligned}
    \label{eq:out_of_sample_reward}
\end{equation}
The sums run over active sellers only, so the zero score assigned to missing sellers in \eqref{eq:score} affects the long-term score but not this allocation. Furthermore, an active seller receives more than the equal allocation when its score exceeds the average score of the other active sellers, and less otherwise. Missing sellers receive zero. When only one seller is active, no relative performance comparison is possible and the full out-of-sample allocation is assigned to that seller.
}

\subsubsection{Properties} \label{properties}
In our setting, the pay-off allocation function must satisfy key economic properties to incentivize participation, encourage truthful forecasts, and ensure consistent rewards. We formally define these properties, and sketch the proof of their validity, in the following. \textcolor{black}{These properties hold for sellers with distinct identities who act independently. Participation of one entity through multiple identities and collusion among sellers are not considered.}

\textbf{Budget Balance}: This property guarantees that the market operator redistributes the exact amount of utility among the sellers. \textcolor{black}{Both allocations use the equal-stake rule of \cite[Theorem~2]{Lambert2008}, which is budget balanced for any vector of scores in $[0,1]$, so that $\sum_i r_{i,t}^{\mathrm{is}(\tau)} = \sum_i r_{i,t}^{\mathrm{oos}(\tau)} = 1$. The total payment is then
\begin{equation}
    \begin{split}
        \sum_i R_{i,t}
        &= \sum_{\tau \in \mathbf{m}} U_t^{(\tau)}\left[\delta \sum_i r_{i,t}^{\mathrm{is}(\tau)} + (1-\delta)\sum_i r_{i,t}^{\mathrm{oos}(\tau)}\right] \\
        &= \sum_{\tau \in \mathbf{m}} U_t^{(\tau)} = U_t.
    \end{split}
\end{equation}
Thus, the sum of distributed payments equals the total utility generated. The result does not depend on the submitted forecasts, the realization, the past scores, or which sellers are active.
}

\textcolor{black}{
\textbf{Symmetry (interchangeability)}: Two sellers who are active in the same sessions and submit identical forecasts in each session up to time \(t\) receive identical rewards. Their scores in \eqref{eq:score} are therefore equal at every session, so their long-term scores also coincide: $\bar{S}_{i,t}^{(\tau)} = \bar{S}_{j,t}^{(\tau)}$. Under the equal-stake rule, each seller’s reward depends on its own score and the sum of the scores of the other active sellers. Since the sellers’ scores are equal, these sums are also equal. Thus, the sellers receive equal rewards under both \eqref{eq:in_sample_reward} and \eqref{eq:out_of_sample_reward}, giving $R_{i,t} = R_{j,t}$.
}

\textbf{No reward without participation}: Sellers who do not participate should receive no reward. Indeed, if seller $i$ does not submit a forecast at time $t$, we set $\alpha_{i,t} = 1$. Based on the reward construction in \eqref{eq:in_sample_reward} and \eqref{eq:out_of_sample_reward}, this explicitly forces the reward to zero, satisfying the property for missing forecasts.

\textcolor{black}{
\textbf{Individual rationality}: This guarantees that no seller is penalized for participating (i.e., pay-offs are non-negative). Since all scores lie in $[0,1]$, the worst case for seller $i$ in \eqref{eq:in_sample_reward} and \eqref{eq:out_of_sample_reward} is a score of $0$ while all other active sellers score $1$, which gives a reward of exactly zero. In every other case the reward is positive, and inactive sellers receive zero. Hence $R_{i,t} \geq 0$ for all $i$ and $t$.
}

\textcolor{black}{
\textbf{Truthfulness (incentive compatibility)}: Assuming that sellers are risk-neutral and that each seller's report and participation are independent of those of the other sellers, a seller maximizes its expected reward by reporting the quantiles of its own predictive distribution. For $N_t \geq 2$, the out-of-sample allocation in \eqref{eq:out_of_sample_reward} is the equal-stake rule of \cite{Lambert2008}, which is truthful for any strictly proper score bounded in $[0,1]$. The score in \eqref{eq:score} satisfies both conditions when the $\tau$-quantile is unique, since the pinball loss is then uniquely minimized in expectation by the true $\tau$-quantile, and the normalization is strictly increasing on $[\underline{y}, \overline{y}]$, to which all reports are clipped. The in-sample allocation in \eqref{eq:in_sample_reward} is increasing and affine in $\bar{S}_{i,t}^{(\tau)}$, which by \eqref{eq:long_term_score} increases with $sc_{i,t}^{(\tau)}$ in the current and all later sessions. Truthful reporting therefore also maximizes the expected current and future in-sample rewards. When $N_t = 1$, the current reward does not depend on the report, but the score still enters the long-term score, so truthful reporting remains optimal.
}

\section{Application and Case-studies} \label{sec:evaluation}
To demonstrate the proposed market and methods, we begin by evaluating the forecasting combination algorithms and pay-off allocation across several examples, starting with two synthetic test cases and concluding with a real-world forecasting scenario. These case studies are, of course, simplified versions of what would be implemented in real-world applications. \textcolor{black}{In particular, the sellers are not strategic agents, and missing submissions are generated at random. The case studies therefore evaluate the forecast combination and the pay-off allocation, while the economic properties of the mechanism follow from the analytical results in Section~\ref{properties}, so they are not empirically tested.}

\subsection{Synthetic Test Cases}
To test the methods proposed in the previous sections, we generated two different environments. First, we implement a time-invariant process that shows the efficacy of the methods proposed and their convergence. Then, a time-varying process is proposed to showcase the ability of the methods to adapt to changes in the dynamics of the environment.

In both scenarios, we consider a single client and three sellers. Each forecaster is modeled using a Normal distribution, and the realizations $Y_t$ are generated as a combination of the sellers' distributions. \textcolor{black}{Since the sellers' forecasts and the realizations are unbounded, we fix a priori the bounds $[\underline{y}, \overline{y}] = [-3, 5]$, which cover the generating distributions with a margin of  4 standard deviations around the mean realization, and clip values outside them. Clipping affects about $0.02\%$ of the realizations and $0.03\%$ of the forecasts}. We refer to the standard linear regression model using quantile regression as \textit{QR}, and as \textit{RQR} to the robust implementation. To evaluate performance, we performed a Monte Carlo simulation consisting of 1000 independent experiments, with $T = 20\,000$. For all scenarios, the learning rate of the aggregation method was set to $\eta = 0.1$, the pay-off allocation parameter to $\delta = 0.7$, and the forgetting factor to $\lambda = 0.999$. \textcolor{black}{The value of $\delta$ gives more weight to long-term consistency than to instantaneous performance, and $\lambda = 0.999$ corresponds to an effective window of $n_\lambda = 1000$ steps}. \textcolor{black}{Whenever missing forecasts are considered, missingness is generated independently across sellers and sessions. In each experiment, each seller forecast is independently marked as unavailable according to the defined missingness rate, subject to the constraint that at least one forecast remains available at every session.}

\subsubsection{Time-invariant case}
The primary goal of this scenario is to verify that the proposed methods work correctly and converge to optimal weights over time. Let $\mu_{i, t}$ and $\sigma_{i, t}$ denote the mean and standard deviation of the $i$-th seller at time $t$. The seller's distribution is defined as follows
    \begin{equation}
        f_{i, t} \sim \mathcal{N}(\mu_{i, t}; \sigma_{i, t})
    \end{equation}
where $\mu_{i, t} = C_i + \alpha \epsilon_{i, t}$, with $C_i$ constant, $\epsilon_{i, t} \sim \mathcal{N}(0,1)$. In our setup, we consider three sellers having $C_1 = 0$, $C_2 = 1$ and $C_3 = 2$, $\alpha = 0.5$, and $\sigma_{1, t} = \sigma_{2, t} = \sigma_{3, t} = 1$.

Finally, the realizations are generated as follows
    \begin{equation} \label{eq:realizations}
        Y_t \sim \mathcal{N}(\mu_t; \sigma_t)
    \end{equation}
where $\mu_t = \sum_{i=1}^n w_i \mu_{i, t}$, $\sigma_t = \sum_{i=1}^n w_i \sigma_{i, t}$, and $\textbf{w}$ is the vector of weights to be learned. In our case, the weights are set to $\textbf{w} = [0.1, 0.6, 0.3]$.

\begin{figure}[htb]
        \centering
        \includegraphics[width=\columnwidth]{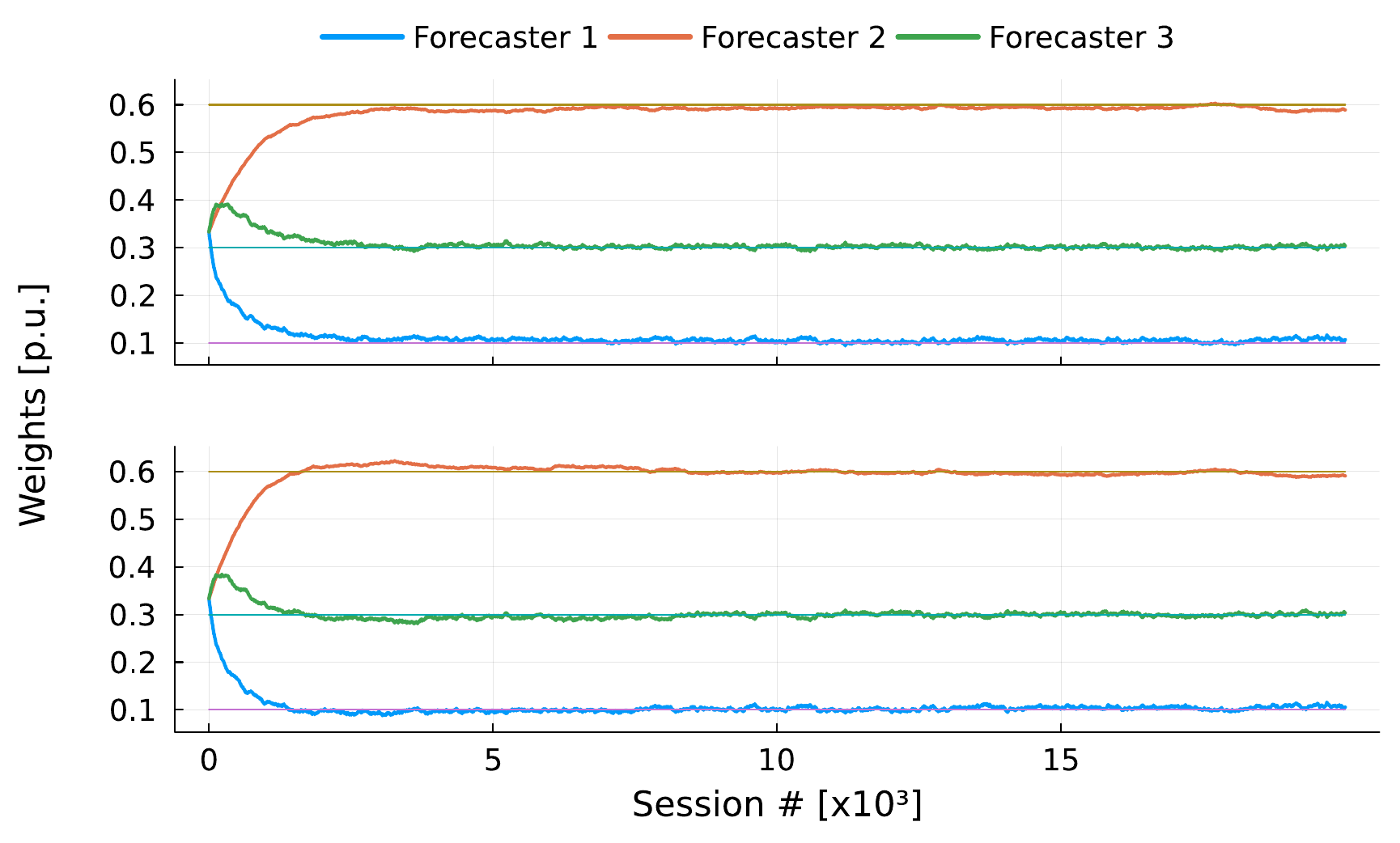}
        \caption{Convergence of estimated weights for QR (top) and RQR (bottom) with $k=1$ and $\tau=0.5$.}
        \label{fig:plot_weights_q50}
    \end{figure}

Fig. \ref{fig:plot_weights_q50} illustrates the convergence behavior of both proposed methods (QR top, RQR bottom) with time horizon $k=1$ and quantile level $\tau = 0.5$. In the RQR case, forecasters are randomly missing with probability $5\%$. As expected, both algorithms converge over time toward the optimal weights combination.

\begin{figure}[htb]
        \centering
        \includegraphics[width=0.9\columnwidth]{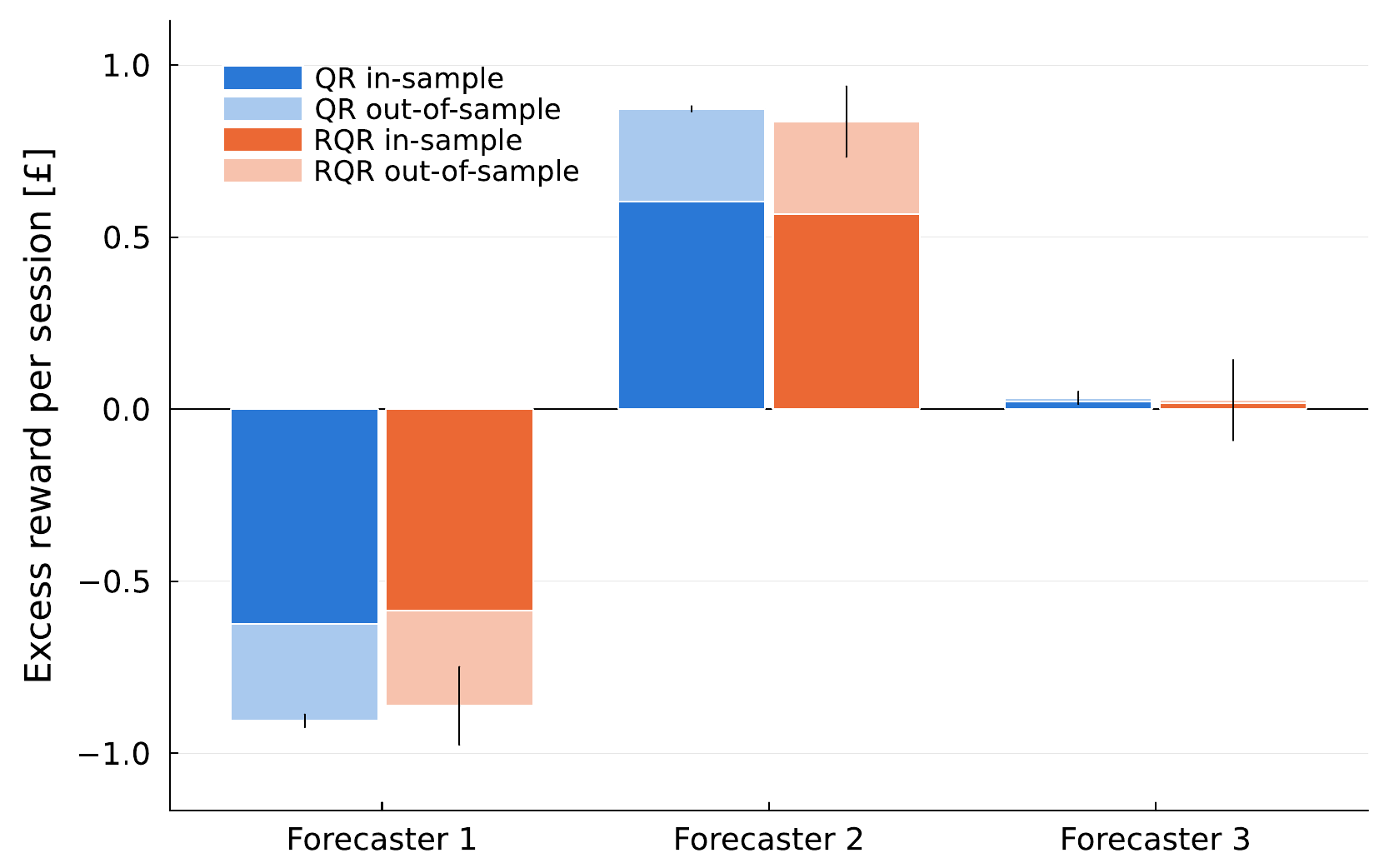}
        \caption{\textcolor{black}{Average reward per session in excess of the equal share for QR and RQR (5\% missingness), split into in-sample (dark) and out-of-sample (light) components. Error bars show $\pm1$ standard deviation across runs.}}
        \label{fig:plot_rewards_all_q}
    \end{figure}

\textcolor{black}{Fig.~\ref{fig:plot_rewards_all_q} shows the average reward per session of each seller in excess of the equal share $U_t/N \approx \pounds 33.3$. We consider three quantile levels, $\mathbf{m} = \{0.1, 0.5, 0.9\}$, with a total reward of $U_t = \pounds 100$ per session equally distributed across levels. This value of $U_t$ is illustrative, and any other fixed amount agreed when the market is opened could be used. The ranking follows the weights used to generate the realizations: Forecaster~2 receives about £0.9 per session above the equal share, Forecaster~1 about £0.9 below, and Forecaster~3 roughly the equal share. The excesses sum to zero, as required by budget balance, and the in-sample component accounts for about 70\% of each, in line with $\delta = 0.7$. RQR yields almost the same allocation as QR, with larger variability across runs due to the random missingness. The differences are small, about 3\% of the equal share, since under the equal-stake rule the excess reward is proportional to the difference between normalized scores, which all lie close to one.}

\subsubsection{Time-varying case}
In this setting, we simulate a dynamic environment in which the combination of weights evolves over time. The goal is to demonstrate that the proposed methods can detect these changes and adapt accordingly. This property is crucial, as real-world applications are inherently dynamic and subject to temporal variation. To do so, we introduce a periodic coefficient defined as
    \begin{equation}
        \beta_t = \dfrac{1}{2} \left( 1 + \sin\!\left(\dfrac{2\pi t}{T}\right) \right) , \quad \beta_t \in [0, 1]
    \end{equation}
Using this coefficient, we define the target weight vector as
    \begin{equation}
        \textbf{w}_t^{\text{target}} = (1 - \beta_t) \textbf{w}^{(1)} + \beta_t \textbf{w}^{(2)}
    \end{equation}
where $\textbf{w}^{(1)} = [0.1, 0.6, 0.3]$ and $\textbf{w}^{(2)} = [0.4, 0.2, 0.4]$ are two different weight combinations. The actual weights used at time $t$ are then updated recursively
    \begin{equation}
        \textbf{w}_t = \rho \textbf{w}_{t-1} + (1 - \rho) \textbf{w}_t^{\text{target}}
    \end{equation}
with $\rho = 0.999$. Finally, the realizations are generated according to (\ref{eq:realizations}).

\begin{figure}[htb]
        \centering
        \includegraphics[width=0.95\columnwidth]{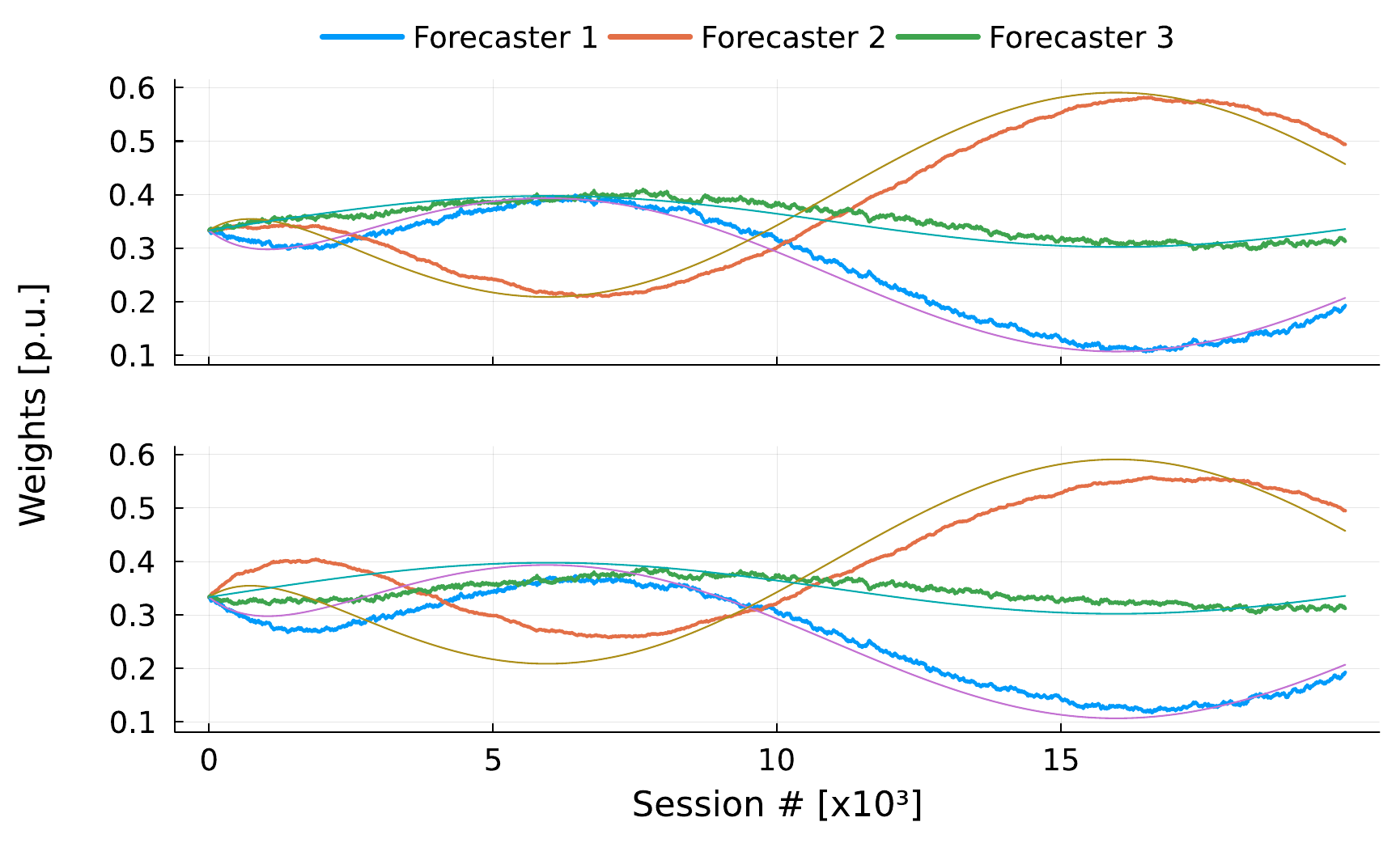}
        \caption{Convergence of estimated weights for QR (top) and RQR (bottom) with $k=1$ and $\tau=0.5$ in a time-varying scenario. \vspace{-2mm}}
        \label{fig:plot_weights_varying_case}
    \end{figure}

The results shown in Fig. \ref{fig:plot_weights_varying_case} report the evolution of the estimated weights in a dynamic scenario. Both methods are able to adapt to the evolving environment, with the estimated weights following the underlying periodic pattern. As expected, convergence is not exact, but the overall dynamics are well captured.
Also in this case the quantile level considered is $\tau = 0.5$, and for RQR (bottom) the missing rate is $5\%$.

\subsection{Performance with Varying Missingness}
In this section, we compare the adaptive QR methods against two benchmark imputation strategies: mean imputation and last-value imputation. The evaluation focuses on the model’s tracking performance and how it is affected by different rates of missingness over time. All results are presented for the quantile level $\tau = 0.1$.

\begin{figure}[htb]
        \centering
        \includegraphics[width=.95\columnwidth]{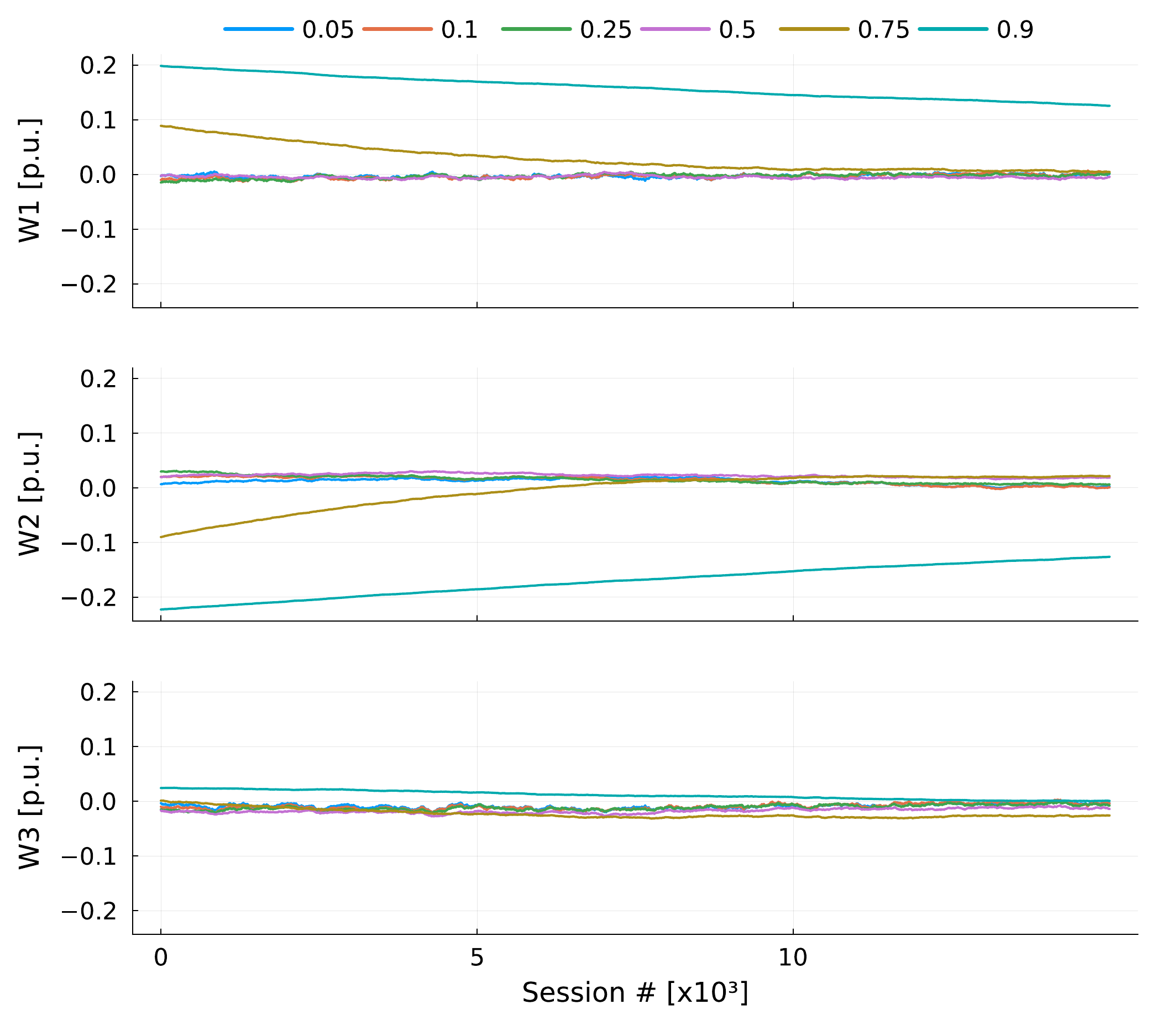}
        \caption{Bias with varying missingness rate ranging from 5\% to 90\%. \vspace{-2mm}}
        \label{fig:bias_varying_missingness}
    \end{figure}

\textcolor{black}{We consider the time-invariant setting, with true weights $\mathbf{w} = [0.1, 0.6, 0.3]$. For each method, we compute the bias and variance of the estimated weights, excluding the first $5000$ steps so that only the results after convergence are considered. Table~\ref{table:metrics} reports both metrics for missingness rates from $5\%$ to $90\%$, with the constraint that at least one seller is always present, and Fig.~\ref{fig:bias_varying_missingness} shows the bias of RQR as a function of the missingness rate.}

\textcolor{black}{RQR has the lowest variance for all weights and missingness rates, and the gap with the imputation methods widens as missingness increases. Up to $50\%$, the bias of RQR is small, about $1$--$3\%$ of the true weights at $5\%$ missingness, and grows moderately with the missingness rate, since the model has less information to learn the combination and the linear correction. Mean-impute has a larger and growing bias for $w_1$, but a lower bias for $w_2$ and $w_3$. Last-value impute has the largest bias up to $75\%$, which grows steadily as outdated forecasts are reused for longer periods. At $75\%$ and $90\%$, the performance of RQR deteriorates. At $90\%$, the biases for $w_1$ and $w_2$ are large and of opposite sign, exceeding those of mean-impute and, for $w_1$, also that of last-value impute. A possible explanation is that, when two sellers are rarely available at the same time, the model has little information to separate their contributions.}

    \begin{table*}[htb]
        \caption{\textcolor{black}{Bias and variance of the estimated weights for the proposed RQR and the two benchmark models under varying missingness rates (all values are in units of $10^{-3}$, reported as mean $\pm$ standard deviation across runs).}}
        \label{table:metrics}
        \centering
        \footnotesize
        \renewcommand{\arraystretch}{0.94}
        \begin{tabular}{cccccccc}
            \toprule
            & & \multicolumn{2}{c}{$w_1$} & \multicolumn{2}{c}{$w_2$} & \multicolumn{2}{c}{$w_3$}\\
            \cmidrule(r){3-8}
            Missingness & Model & bias & var & bias & var & bias & var \\
            \midrule
            \multirow{3}{*}{5\%}  & RQR         & $-2.8 \pm 3.2$  & $1.91 \pm 0.16$ & $11.7 \pm 5.6$   & $1.87 \pm 0.23$ & $-8.9 \pm 4.2$  & $2.08 \pm 0.21$ \\
                                  & Last-value Impute & $16.7 \pm 3.7$  & $2.06 \pm 0.19$ & $-20.1 \pm 6.6$  & $2.04 \pm 0.21$ & $3.4 \pm 4.6$   & $2.13 \pm 0.22$ \\
                                  & Mean-impute & $8.6 \pm 3.7$   & $2.03 \pm 0.19$ & $-7.9 \pm 6.3$   & $2.02 \pm 0.22$ & $-0.7 \pm 4.4$  & $2.12 \pm 0.22$ \\
            \midrule
            \multirow{3}{*}{10\%} & RQR         & $-3.8 \pm 3.9$  & $1.85 \pm 0.17$ & $13.5 \pm 7.2$   & $1.75 \pm 0.18$ & $-9.7 \pm 5.0$  & $2.08 \pm 0.21$ \\
                                  & Last-value Impute & $26.3 \pm 3.8$  & $2.10 \pm 0.19$ & $-32.0 \pm 7.2$  & $2.08 \pm 0.20$ & $5.7 \pm 5.0$   & $2.15 \pm 0.23$ \\
                                  & Mean-impute & $10.4 \pm 4.3$  & $2.01 \pm 0.18$ & $-8.1 \pm 7.3$   & $1.94 \pm 0.19$ & $-2.3 \pm 4.8$  & $2.11 \pm 0.22$ \\
            \midrule
            \multirow{3}{*}{25\%} & RQR         & $-3.5 \pm 4.1$  & $1.73 \pm 0.16$ & $15.0 \pm 6.8$   & $1.49 \pm 0.17$ & $-11.5 \pm 4.6$ & $2.06 \pm 0.21$ \\
                                  & Last-value Impute & $54.7 \pm 3.6$  & $2.18 \pm 0.21$ & $-69.2 \pm 6.0$  & $2.13 \pm 0.17$ & $14.5 \pm 4.6$  & $2.17 \pm 0.23$ \\
                                  & Mean-impute & $14.9 \pm 4.9$  & $2.06 \pm 0.19$ & $-8.9 \pm 9.1$   & $1.95 \pm 0.16$ & $-6.0 \pm 5.5$  & $2.10 \pm 0.22$ \\
            \midrule
            \multirow{3}{*}{50\%} & RQR         & $-5.0 \pm 2.5$  & $1.57 \pm 0.14$ & $22.6 \pm 3.4$   & $1.19 \pm 0.15$ & $-17.6 \pm 4.1$ & $2.01 \pm 0.24$ \\
                                  & Last-value Impute & $97.7 \pm 3.3$  & $2.34 \pm 0.21$ & $-126.9 \pm 4.4$ & $2.40 \pm 0.27$ & $29.2 \pm 4.0$  & $2.26 \pm 0.23$ \\
                                  & Mean-impute & $26.4 \pm 8.6$  & $2.11 \pm 0.19$ & $-17.9 \pm 16.7$ & $1.98 \pm 0.20$ & $-8.5 \pm 8.8$  & $2.10 \pm 0.22$ \\
            \midrule
            \multirow{3}{*}{75\%} & RQR & $28.4 \pm 23.8$ & $1.58 \pm 0.18$ & $-5.6 \pm 31.6$  & $1.16 \pm 0.16$ & $-22.9 \pm 8.1$ & $1.77 \pm 0.19$ \\
            & Last-value Impute & $128.3 \pm 3.1$ & $2.45 \pm 0.24$ & $-168.4 \pm 2.6$ & $2.44 \pm 0.21$ & $40.1 \pm 3.2$  & $2.36 \pm 0.24$ \\
            & Mean-impute & $39.9 \pm 12.5$ & $2.11 \pm 0.20$ & $-34.4 \pm 24.9$ & $2.00 \pm 0.15$ & $-5.5 \pm 13.0$ & $2.19 \pm 0.21$ \\
            \midrule
            \multirow{3}{*}{90\%} & RQR         & $158.9 \pm 20.5$ & $0.98 \pm 0.25$ & $-170.4 \pm 28.1$ & $0.66 \pm 0.15$ & $11.5 \pm 7.7$  & $1.12 \pm 0.20$ \\
                                  & Last-value Impute & $137.9 \pm 3.1$  & $2.42 \pm 0.26$ & $-181.0 \pm 3.6$  & $2.46 \pm 0.35$ & $43.2 \pm 3.5$  & $2.42 \pm 0.24$ \\
                                  & Mean-impute & $47.6 \pm 14.7$  & $2.11 \pm 0.19$ & $-45.6 \pm 29.0$  & $1.92 \pm 0.13$ & $-2.1 \pm 14.8$ & $2.16 \pm 0.21$ \\
            \bottomrule
        \end{tabular}
    \end{table*}

\subsection{Real-world Forecasting Problem}
We demonstrate the proposed market framework through a wind energy forecasting case study. The analysis focuses on day-ahead probabilistic forecasting of the aggregate power production of the Belgian offshore wind fleet. The power production data are obtained from the open data portal of the Belgian transmission system operator, Elia Group \cite{eliawebsite}. The fleet has a nominal installed capacity of 2,262.1 MW, and the dataset covers the period from 2024-01-01 to 2025-12-31 at a 15 min resolution. As sellers, we consider nine forecasting models. We use forecasts from three Numerical Weather Prediction (NWP) models, namely (i) the High-Resolution Forecast (HRES) from the European Centre for Medium-Range Weather Forecasts (ECMWF) \cite{ecmwf_ifs}, (ii) the Global Forecast System (GFS) from the National Oceanic and Atmospheric Administration (NOAA) \cite{noaa_gfs}, and (iii) the ICON-EU from Deutscher Wetterdienst (DWD) \cite{dwd_icon}. For each NWP dataset, we consider a grid surrounding the fleet location. These spatial features are subsequently transformed using Principal Component Analysis (PCA) in order to obtain a reduced set of uncorrelated predictors. The resulting pre-processed features are then used to train three different probabilistic forecasting models for each NWP source. Specifically, we employ (i) Extreme Gradient Boosting (XGB), (ii) Quantile Regression Forests (QRF), and (iii) a Multi-Layer Perceptron (MLP). Since three models are fitted for each of the three NWP datasets, this results in a total of nine probabilistic forecasts that are subsequently considered as input to the market. 
It is worth emphasizing that the primary objective of this study is not to develop the most accurate individual forecasting models, but rather to demonstrate the effectiveness of the proposed market framework.

Moreover, the individual forecasting models were trained exclusively on data from 2024. Then, validation and testing of the online prediction market were performed on data from 2025. To improve convergence and reduce gradient variance when adapting market weights across multiple lead times, we implemented a mini-batch strategy for the online gradient updates, averaging the loss across each batch. Specifically, the learning rate and batch size of the online prediction market were selected on a validation set made up of the first two months of 2025 (January and February). We conducted a grid search over the learning rates $\eta \in \{0.2, 0.1, 0.05, 0.01, 0.005\}$ and batch percentages $b \in \{0.05, 0.1, 0.2, 0.5, 1.0\}$. The optimal hyperparameters for the market mechanism were identified as 0.1 for both the learning rate and the batch size percentage. \textcolor{black}{As in the synthetic case studies, the client pays $U_t = \pounds 100$ per daily session, split equally across the three quantile levels, and the pay-off parameter is set to $\delta = 0.7$. The forgetting factor is set to $\lambda = 0.989$, which corresponds to an effective window of three months ($n_\lambda = 90$ sessions).} Final testing was then performed on the remaining ten months of 2025, where the initial two-month validation period was utilized as a burn-in period to initialize the online mechanism.

To assess the robustness of the results, we conduct a Monte Carlo simulation consisting of 100 independent experiments. The reported performance metrics correspond to the average results over these runs.

\begin{table}[!h]
    \centering
    \caption{\textcolor{black}{Quantile loss (in MW) for different quantile levels and methods. For each quantile level, underlined values denote the best individual model and bold values the best method overall.}}
    \begin{tabular}{llccc}
        \toprule
         & \multirow{2}{*}{Model} & \multicolumn{3}{c}{Quantile} \\ 
        \cmidrule(l){3-5}
         & & 0.1 & 0.5 & 0.9 \\
        \midrule
        \multirow{3}{*}{ECMWF} 
         & QRF & 37.76 & 86.68 & 45.78 \\
         & MLP & 37.18 & 85.45 & \underline{41.05} \\
         & XGB & \underline{36.19} & \underline{82.38} & 43.92 \\
        \addlinespace
        \multirow{3}{*}{NOAA}   
         & QRF & 41.11 & 100.42 & 52.24 \\
         & MLP & 40.13 & 96.98 & 49.14 \\
         & XGB & 38.66 & 95.96 & 50.79 \\
        \addlinespace
        \multirow{3}{*}{DWD}   
         & QRF & 40.14 & 96.40 & 49.59 \\
         & MLP & 38.13 & 90.36 & 44.48 \\
         & XGB & 38.43 & 91.55 & 48.12 \\
        \addlinespace
        \multirow{7}{*}{Combination} 
         & QR         & \textbf{34.15} & \textbf{77.93} & \textbf{37.38} \\
         & RQR - 5\%  & 34.21 & 78.20 & 37.61 \\
         & RQR - 10\% & 34.28 & 78.43 & 37.80 \\
         & RQR - 20\% & 34.44 & 78.88 & 38.28 \\
         & RQR - 25\% & 34.48 & 79.18 & 38.49 \\
         & RQR - 50\% & 35.17 & 81.06 & 39.97 \\
         & RQR - 75\% & 36.65 & 85.89 & 43.11 \\
         & RQR - 90\% & 38.16 & 90.37 & 46.07 \\
        \bottomrule
    \end{tabular}
    \label{res:quantile_loss_models}
    \vspace{-5mm}
\end{table}

\textcolor{black}{Table~\ref{res:quantile_loss_models} shows the quantile loss achieved by all considered methods across the different quantile levels. Models trained using weather forecasts from ECMWF generally perform better than those based on NOAA and DWD. Among the combination methods, QR performs best and serves as an oracle benchmark, since it assumes that all forecasts are available. Combining forecasts based on different information sources therefore improves probabilistic accuracy compared with any individual model.}

\textcolor{black}{The main objective of this work is to design a method that remains effective when forecasts are missing. Up to a missingness rate of $25\%$, the quantile loss of RQR is within $3\%$ of the oracle QR benchmark. At $50\%$, the increase is between $3\%$ and $7\%$, depending on the quantile level, and RQR still outperforms the best individual model. From $75\%$, its loss exceeds that of the best individual model, since the combination relies on too few available forecasts to compensate for the missing ones.}

\begin{figure}[htb]
    \centering
    \includegraphics[width=\columnwidth]{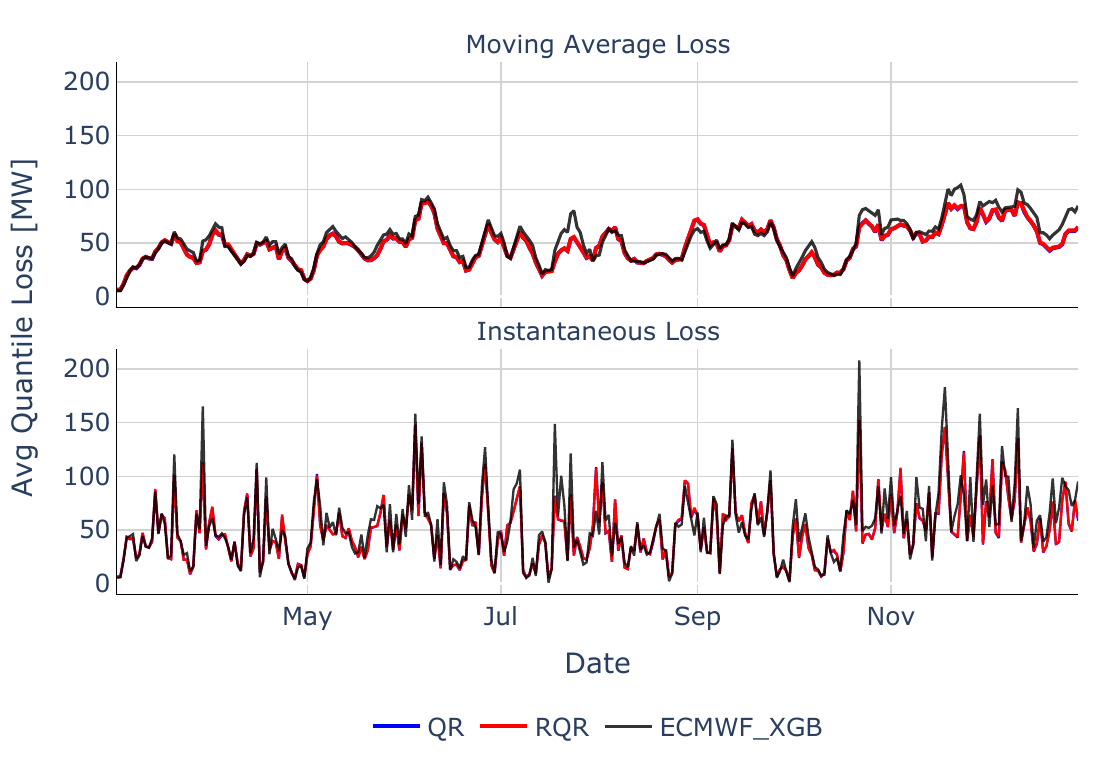}
    \caption{Moving average and instantaneous loss of the best-performing seller as well as the online combination methods.}
    \label{fig:plot_losses}
\end{figure}

Fig.~\ref{fig:plot_losses} illustrates the instantaneous loss (bottom panel) and seven-day moving average loss (top panel) for the most accurate individual seller (ECMWF-XGB) alongside the combination methods. While the instantaneous loss highlights the high daily volatility of the predictions, it shows that the oracle (QR) and RQR (with 5\% missingness) generally mitigate the most severe error spikes compared to the single seller. 
The advantages of these adaptive combination techniques become particularly evident when examining the moving average loss. Here, RQR demonstrates robust predictive capability, consistently maintaining a lower moving average loss than the single best forecaster. It becomes clear that adaptive techniques prove highly beneficial in instances where one or more sellers underperform; this is notably visible after November, where the moving average loss of the single seller spikes significantly above the combination methods.

\begin{figure}[htb]
    \centering
    \includegraphics[width=\columnwidth]{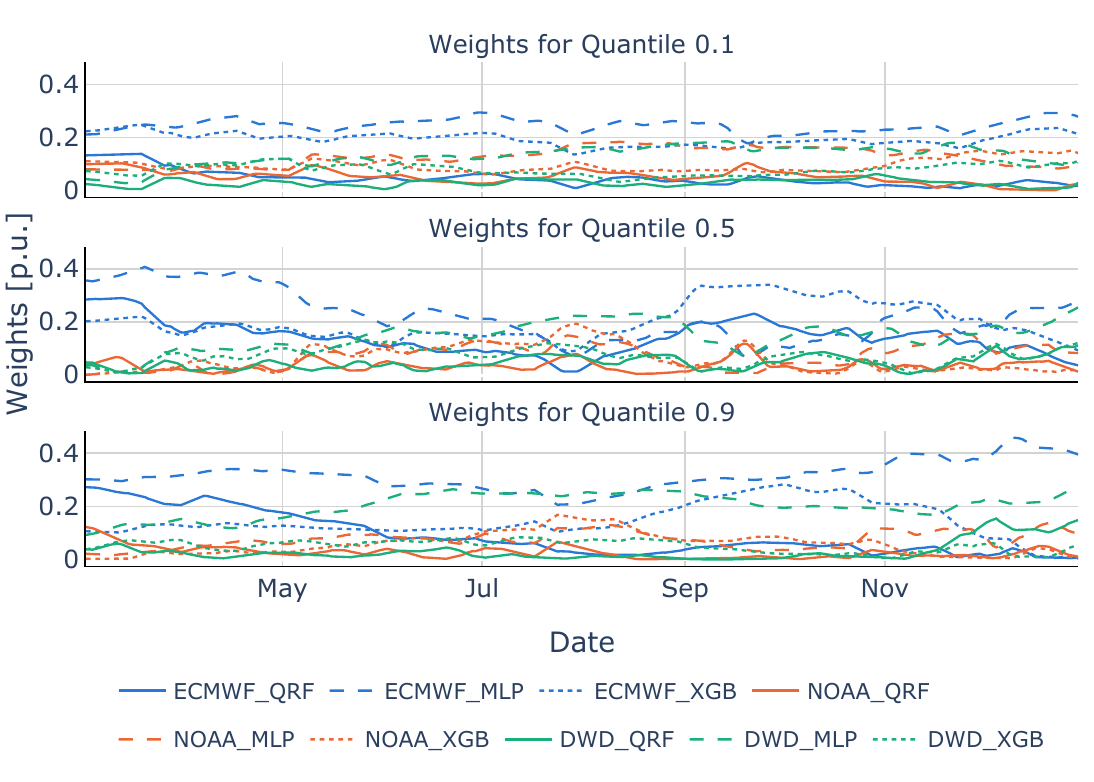}
    \caption{\textcolor{black}{Seven-day moving average of the combination weights for each quantile level over the test period for RQR with 5\% missingness.}}
    \label{fig:plot_weights}
\end{figure}

Fig.~\ref{fig:plot_weights} displays the seven-day moving average of the combination weights for each quantile level over the test period. \textcolor{black}{Overall, the largest weights are assigned to ECMWF-MLP and ECMWF-XGB across all quantile levels, which is consistent with the results in Table~\ref{res:quantile_loss_models}.
Interestingly, the DWD-MLP seller exhibits a distinct time-varying behavior. Its weight is small at the start of the test period but grows steadily over the first months for all quantile levels. Between July and August, it attains the highest weight for the median (0.5) and is comparable to ECMWF-MLP for the upper quantile (0.9). This occurs despite its overall performance being less competitive when evaluated across the entire test set.}

\begin{figure}[htb]
    \centering
    \includegraphics[width=0.9\columnwidth]{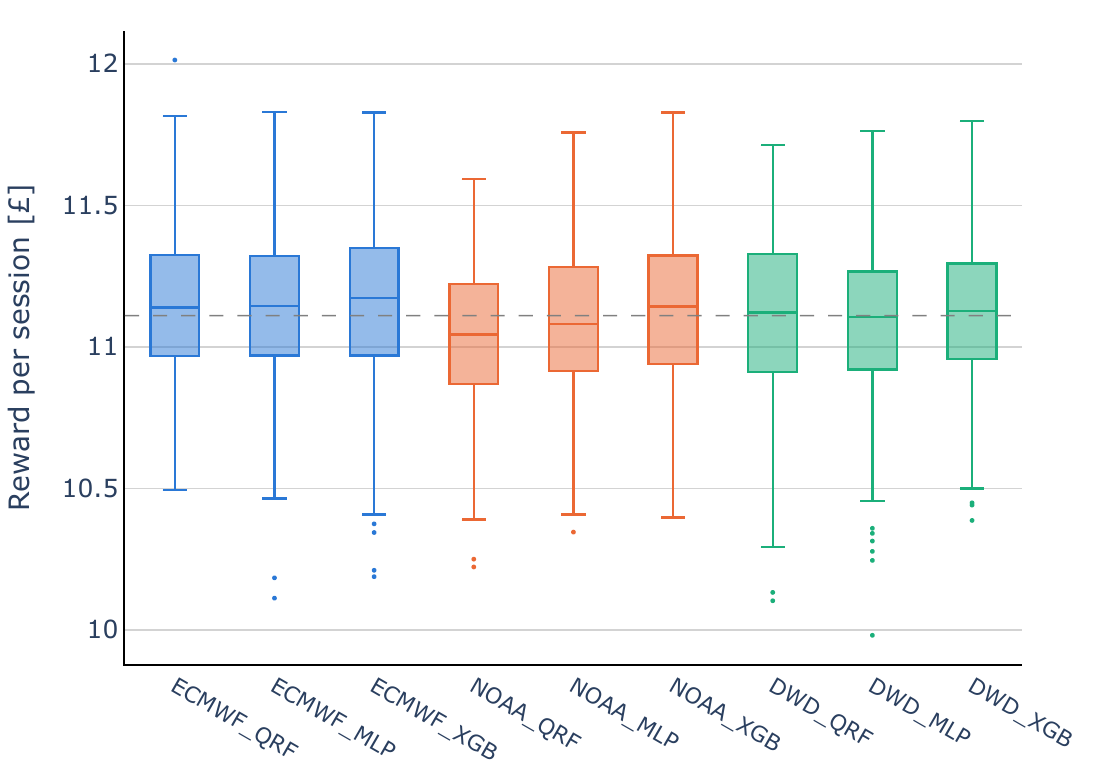}
    \caption{\textcolor{black}{Distribution of rewards per session for all market participants considering the RQR method with 5\% missingness. The dashed line indicates the equal share of the total reward.}}
    \label{fig:plot_rewards}
\end{figure}

\begin{table*}[t]
    \centering
    \caption{\textcolor{black}{Monthly rewards (in £) of each seller from March to December 2025, using RQR method with 5\% missingness. Bold values denote the highest reward in each month.}}
    \renewcommand{\arraystretch}{0.9}
    \setlength{\tabcolsep}{4pt}
    \begin{tabular*}{\textwidth}{@{\extracolsep{\fill}} llcccccccccc @{}}
        \toprule
         & Model & Mar & Apr & May & Jun & Jul & Aug & Sep & Oct & Nov & Dec \\ 
        \midrule
        \multirow{3}{*}{ECMWF} 
         & QRF & 334.10 & 333.38 & 344.32 & 334.17 & 343.89 & \textbf{346.44} & 337.09 & \textbf{348.14} & 334.23 & 332.86 \\
         & MLP & 337.11 & 334.49 & 346.72 & 334.72 & 345.01 & 344.64 & 332.96 & 345.43 & \textbf{336.06} & 332.14 \\
         & XGB & \textbf{337.23} & \textbf{336.48} & 345.91 & 334.02 & 345.23 & 345.67 & \textbf{337.59} & 345.14 & 329.52 & 332.51 \\
        \addlinespace
        \multirow{3}{*}{NOAA}   
         & QRF & 331.16 & 332.09 & 339.89 & 331.30 & 341.44 & 343.75 & 328.76 & 341.82 & 332.32 & 332.62 \\
         & MLP & 333.04 & 332.29 & 344.89 & 334.98 & 344.49 & 343.43 & 331.03 & 340.45 & 332.08 & 333.31 \\
         & XGB & 332.40 & 333.79 & \textbf{350.74} & \textbf{335.27} & 345.17 & 345.12 & 333.56 & 344.96 & 333.36 & 328.81 \\
        \addlinespace
        \multirow{3}{*}{DWD}   
         & QRF & 331.59 & 329.29 & 344.53 & 332.23 & 345.69 & 343.03 & 333.49 & 342.67 & 334.56 & \textbf{336.38} \\
         & MLP & 330.42 & 332.46 & 340.93 & 331.68 & 341.83 & 344.79 & 333.94 & 345.66 & 332.36 & 335.64 \\
         & XGB & 332.94 & 335.72 & 342.06 & 331.62 & \textbf{347.24} & 343.13 & 331.59 & 345.73 & 335.51 & 335.73 \\
        \bottomrule
    \end{tabular*}
    \label{res:monthly_rewards_models}
\end{table*}

\textcolor{black}{Fig.~\ref{fig:plot_rewards} illustrates the distribution of rewards per session across all forecasters. Since rewards depend only on forecast accuracy, all forecasters receive amounts close to the equal share of about £11.1, indicated by the dashed line. The models trained on ECMWF data obtain slightly higher median rewards, consistent with their lower quantile loss (see Table~\ref{res:quantile_loss_models}), while NOAA-QRF receives the lowest. The differences remain small, because the normalized scores of all forecasters lie close to one.}

Table~\ref{res:monthly_rewards_models} details the monthly rewards of the nine forecasters. \textcolor{black}{Over the whole period, the three ECMWF models receive the highest rewards, with nearly identical totals, while NOAA-QRF receives the lowest, ranking between sixth and ninth in every month. The differences remain small: within each month, the highest and lowest rewards differ by 1--3\%. The ranking also changes over the year. DWD-MLP ranks between sixth and ninth from March to July, but third in September, October, and December, and in December the three DWD models receive the highest rewards. In contrast, ECMWF-XGB and ECMWF-MLP are among the lowest in December.}

\begin{table}[ht]
    \centering
    \caption{\textcolor{black}{Wall-clock time per market session of the online
    operations, summed over the three quantile levels (mean and 95th
    percentile, in $\mu$s).}}
    \begin{tabular}{lcccc}
        \toprule
         & \multicolumn{2}{c}{QR} & \multicolumn{2}{c}{RQR} \\
        \cmidrule(lr){2-3}\cmidrule(lr){4-5}
        Operation & Mean & P95 & Mean & P95 \\
        \midrule
        Forecast aggregation & 6.7  & 6.8  & 8.6  & 8.4  \\
        Parameter update     & 13.3 & 11.9 & 31.2 & 28.6 \\
        Pay-off calculation  & 6.7  & 6.6  & 15.4 & 14.6 \\
        \bottomrule
    \end{tabular}
    \label{tab:wall_clock}
\end{table}

\textcolor{black}{
Finally, Table~\ref{tab:wall_clock} reports the wall-clock times of the three operations performed online at each market session, measured over all sessions of 100 runs in Julia 1.12 on a single thread of an Apple M4. The first session of each run is excluded, as it includes the compilation time, and the training of the sellers' models is performed offline and is not part of the market operation. The mean can exceed the 95th percentile because of a few slow sessions, e.g., due to garbage collection. A complete market session takes on average about $27\,\mu$s with QR and $55\,\mu$s with RQR, which is negligible compared with the time between sessions, so the mechanism can be run online. RQR is slower mainly in the parameter update, since it also computes the projected displacement in \eqref{eq:pg_step}--\eqref{eq:pg_displacement} and updates the correction matrix $\mathbf{D}^{(\tau)}$.}

\section{Discussion and Limitations} \label{sec:discussions}
\textcolor{black}{
The proposed allocation rewards forecast accuracy, both over the long term and in each session, rather than the information that each seller brings to the combined forecast. A natural alternative would be to reward each seller for its marginal contribution to the combination, for instance through leave-one-out or Shapley values \cite{shapley1953value}, which measure how much the combined forecast improves when the seller's forecast is added. Such rewards, however, conflict with the properties in Section~\ref{properties}. Under the assumptions of \cite{Lambert2008}, a payment rule that is truthful and normal (i.e., an improvement in one seller's score cannot increase the payment of any other seller) must be additively separable in the sellers' individual scores \cite[Lemma~1]{Lambert2008}. Adding budget balance and anonymity leaves only the competitive form used here \cite[Theorem~2]{Lambert2008}. A marginal contribution depends on how a seller's forecast interacts with the others, so it is not separable, and a reward based on it must give up at least one of these properties. A possible direction is the approach of \cite{Pinson2022}, in which the client pays in proportion to the improvement over its own baseline model and Shapley values split this payment among the sellers. Adapting it to our setting would require the client to provide a baseline forecast, and the total reward to depend on the achieved improvement rather than being fixed in advance.}

\textcolor{black}{
The consequence of this choice is visible in the real-world case. ECMWF-QRF receives almost the same total reward as the best seller (Table~\ref{res:monthly_rewards_models}), although its weights for the lower and upper quantiles remain low for most of the year (Fig.~\ref{fig:plot_weights}). The opposite holds for DWD-MLP, which receives large weights but ranks eighth in total reward. In addition, rewards differ little across sellers, since all normalized scores lie close to one, which reduces the incentive to improve forecasts, a problem known as incentive calibration \cite[Section~6.1]{Lambert2008}.
}

\section{Conclusions} \label{sec:conclusions}
The growing availability of data offers unprecedented opportunities to advance forecasting models and improve the integration of renewable energy generation. Yet, competitive interests and concerns over data privacy often prevent stakeholders from sharing information. Here, we proposed a new prediction market platform that enables communication among stakeholders while incentivizing participation through rewards. Our market design addresses several practical challenges and demonstrates how those can be solved. Several challenges remain open. For example, the market can be extended to a fully dynamic setting, where the market allows a dynamic set of participants over time. \textcolor{black}{The behaviour of the adaptive robust method under structured availability patterns, such as prolonged outages of individual sellers, a subset of sellers that is repeatedly unavailable, or missingness that is correlated across sellers, should also be studied. Furthermore, the mechanism could be made robust to strategic behaviour among sellers, such as one entity participating through multiple identities or several sellers coordinating their submissions.} Finally, future work will investigate rewards based on the marginal contribution of each seller, following \cite{Pinson2022}, as well as score functions that increase the spread of the rewards.

\section*{Acknowledgments}
The authors acknowledge Leonardo Pesce for discussion and input related to the mini-batch approach.

\section*{AI Use Declaration}
During the writing process, AI (Claude) was used to rephrase and improve clarity. No text was purely generated by AI.

\bibliographystyle{IEEEtran}
\bibliography{bibliography}

\end{document}